\documentclass[lettersize,journal]{IEEEtran}
\usepackage{amsmath,amsfonts}

\usepackage{algorithm}
\usepackage{array}
\usepackage[caption=false,font=footnotesize,labelfont=rm,textfont=rm]{subfig}
\usepackage{textcomp}
\usepackage{stfloats}
\usepackage{url}
\usepackage{verbatim}
\usepackage{graphicx}
\usepackage[numbers]{natbib}
\usepackage{amssymb}
\usepackage{tabularx}
\usepackage{multirow}
\usepackage{booktabs}
\usepackage{algpseudocode}
\usepackage{pifont}

\begin{document}

\title{Enhancing Layer Interaction Using Key-Correlated Layer Attention}

\author{Jianlong Xiong, ChuanBo Xie, Le Yu, Quansong He, Tao He
\thanks{This work was supported by the National Natural Science Foundation of China under Grant No. 62206189 and the Science and Technology Planed Project of Zigong under Grant No. 2023ZC25 (Corresponding author: Tao He, email: tao\_he@scu.edu.cn).}
\thanks{Jianlong Xiong, Chuanbo Xie, Le Yu, Quansong He, and Tao He are with the College of Computer Science, Sichuan University, Chengdu, 610065, China. ChuanBo Xie is also with the Department of Ultrasound, Zigong Maternal and Child Health Care Hospital, China.}
}

\markboth{Journal of \LaTeX\ Class Files,~Vol.~14, No.~8, August~2021}%
{Shell \MakeLowercase{\textit{et al.}}: A Sample Article Using IEEEtran.cls for IEEE Journals}


\maketitle

\begin{abstract}
Recent advances in network architecture design have introduced layer attention to enhance inter-layer interactions. In such frameworks, each layer queries all preceding layers to establish cross-layer connections. However, layer attention results in quadratic computational complexity with respect to network depth. To mitigate this issue, prior works have proposed Recurrent Layer Attention (RLA) and linear attention mechanisms, which suffer from static information updates and limited long-range cross-layer dependency modeling. To overcome these limitations, we propose Key-Correlated Layer Attention (KCLA), inspired by our observation that Key representations in layer attention exhibit high cosine similarity. KCLA achieves linear computational complexity while preserving dynamic information updates, directly derived from the foundational definition of layer attention. Furthermore, KCLA maintains long-range cross-layer connections and features a fixed spatial complexity, independent of network depth. Empirical evaluations demonstrate that KCLA delivers good performance across diverse tasks, including image recognition, object detection, and medical image segmentation. The code is publicly available at: \url{https://github.com/bgx666/KCLA}.
\end{abstract}


\section{Introduction}
\IEEEPARstart{T}{he} self-attention mechanism \cite{46201} has demonstrated remarkable capability in capturing long-range dependencies. Given an input sequence $\boldsymbol{X} \in \mathbb{R}^{n \times d_{\text{model}}}$ consisting of $n$ tokens with $d_{\text{model}}$-dimensional embeddings, three learnable linear projections generate the Query ($\boldsymbol{Q}$), Key ($\boldsymbol{K}$), and Value ($\boldsymbol{V}$) representations as follows:
\begin{equation}
\boldsymbol{Q} = \boldsymbol{X}\boldsymbol{W}_Q,\quad
\boldsymbol{K} = \boldsymbol{X}\boldsymbol{W}_K,\quad
\boldsymbol{V} = \boldsymbol{X}\boldsymbol{W}_V
\end{equation}
where $\boldsymbol{W}_Q, \boldsymbol{W}_K \in \mathbb{R}^{d_{\text{model}} \times d}$ and $\boldsymbol{W}_V \in \mathbb{R}^{d_{\text{model}} \times d_v}$ are learnable parameters. Each token in the input sequence is projected into respective latent spaces through these parameters. Subsequently, attention scores are computed between Queries and Keys for any two positions in the sequence, and the Values are aggregated based on these attention scores to directly model long-range dependencies. Numerous studies have shown that enhancing inter-layer interactions can significantly improve the representational capacity of neural networks~\cite{7780459, huang2017densely, fan2021sparse, xu2024development}. 
Recent efforts have increasingly focused on incorporating attention mechanisms into the hierarchical structures of Deep Neural Networks (DNNs)~\cite{zhao2021recurrence, fang2023crosslayerretrospectiveretrievinglayer, wang2024strengthening}. The layer attention, first proposed by~\cite{fang2023crosslayerretrospectiveretrievinglayer}, treats each layer output in a DNN as a token in a sequence, upon which the attention mechanism is applied to efficiently capture cross-layer dependencies.

\begin{figure}[htbp]
    \centering
    \subfloat[Layer attention]{%
    \includegraphics[width=\linewidth]{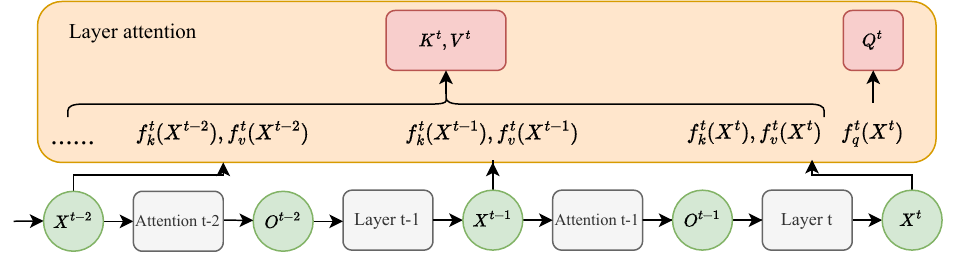}%
    }
    \hfill
    \subfloat[RLA]{%
\includegraphics[width=\linewidth]{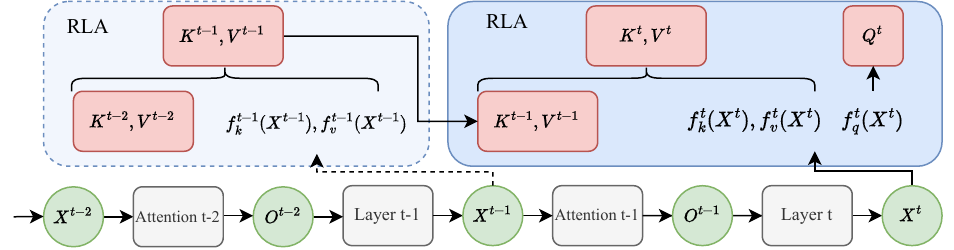}%
    }
    \caption{The details of the standard layer attention module (a) and RLA module (b). The ``Layer" blocks represent the original backbones. The  ``Attention" blocks refer to  modules which can be implemented consistently as either standard layer attention or RLA. The box represented by the dotted line and the dotted arrow specifically illustrate the preceding RLA module and the transformation it performs during the last attention output calculation.} 
    \label{figure:arch of la and rla}
\end{figure}

However, layer attention suffers from three critical issues: (1) the requirement to compute $\boldsymbol{K}$ and $\boldsymbol{V}$ matrices at each layer introduces significant computational overhead, (2) the necessity to cache outputs from previous layers results in memory growth as network depth $D$ increases, (3) query operations incur a computational complexity that scales quadratically with network depth $\mathcal{O}(D^2)$. 

RLA~\cite{fang2023crosslayerretrospectiveretrievinglayer} addressed the first issue by caching the Key and Value representations from all preceding layers. Its multi-head variant is termed Multi-head RLA-Base (MRLA-B). In Fig.~\ref{figure:arch of la and rla}, we provide schematic illustrations of both layer attention and RLA architectures. While this design significantly reduces computational overhead, Dynamic Layer Attention (DLA)~\cite{wang2024strengthening} argued that the simplification in RLA renders it essentially a static mechanism, as the Key and Value representations of earlier layers remain fixed once generated. DLA built a novel but extra dynamic sharing unit to perform dynamic information update, which enlarged the parameter capacity. The second issue existing in layer attention significantly limits the applicability of most linearization methods developed for the vanilla attention mechanism~\cite{wang2020linformer,bolya2022hydra,lingle2023transformer,yang2024gatedlinearattentiontransformers}, as they only focus on solving the problem of the quadratic complexity of the attention mechanism, and still need to cache the output of previous layers. To address the third issue, \cite{fang2023crosslayerretrospectiveretrievinglayer} derived a lightweight MRLA (MRLA-L) via linear approximation of RLA, achieving linear computational complexity and depth-independent space complexity. However, we observe that the linear approximation in MRLA-L causes earlier layer outputs to be exponentially attenuated, making it difficult to establish long-range cross-layer dependencies.


\begin{figure}[htbp]
    \centering
    \subfloat[Between adjacent layers\label{fig:layer-att-ksim-adjacent}]{%
        \includegraphics[width=0.48\linewidth]{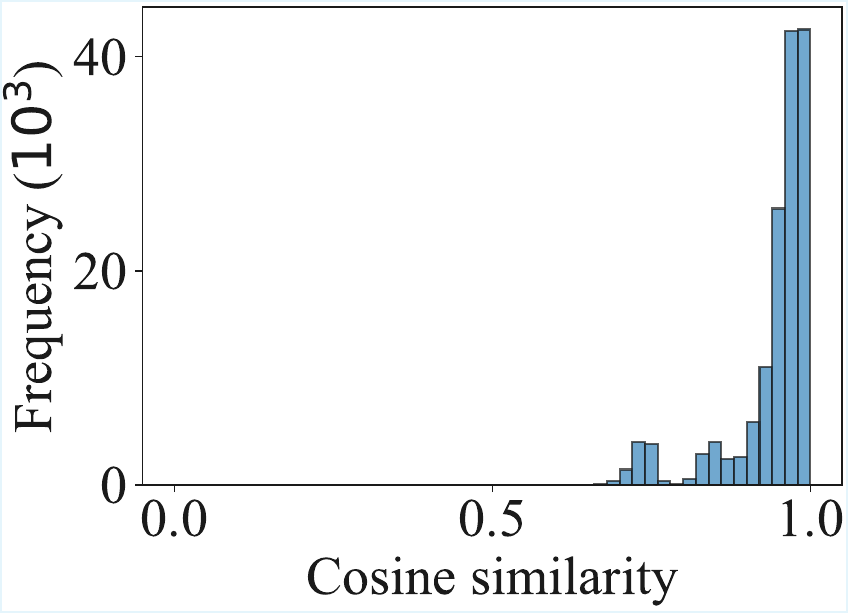}%
    }
    \hfill 
    \subfloat[Between arbitrary layers\label{fig:layer-att-notadjacent}]{%
        \includegraphics[width=0.48\linewidth]{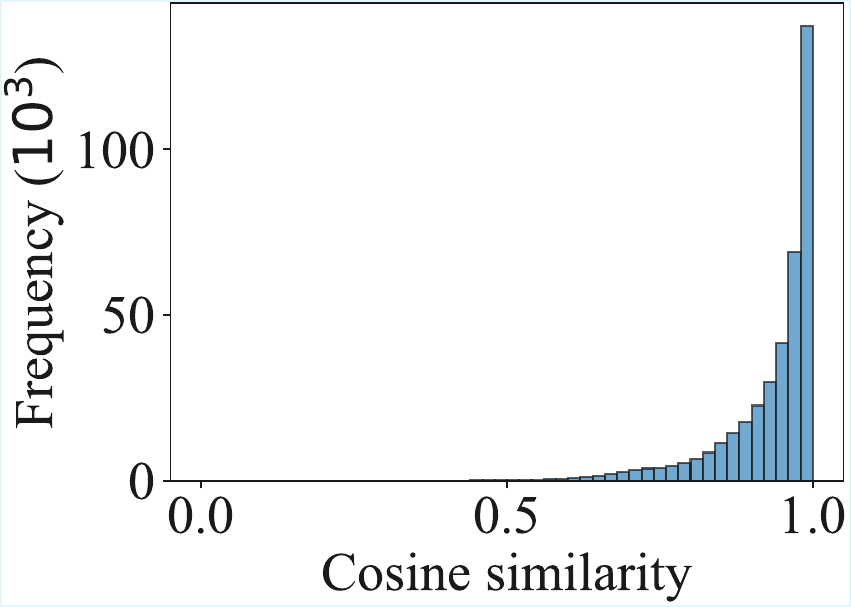}%
    }
    
    \caption{Histogram of cosine similarity distribution for Key representations in standard layer attention. The results in the figure are statistics from the CIFAR-100 validation set.} 
    \label{fig:layer-att-ksim-summary}
\end{figure}

\begin{figure}[htbp]
\centering
\includegraphics[width=0.8\linewidth]{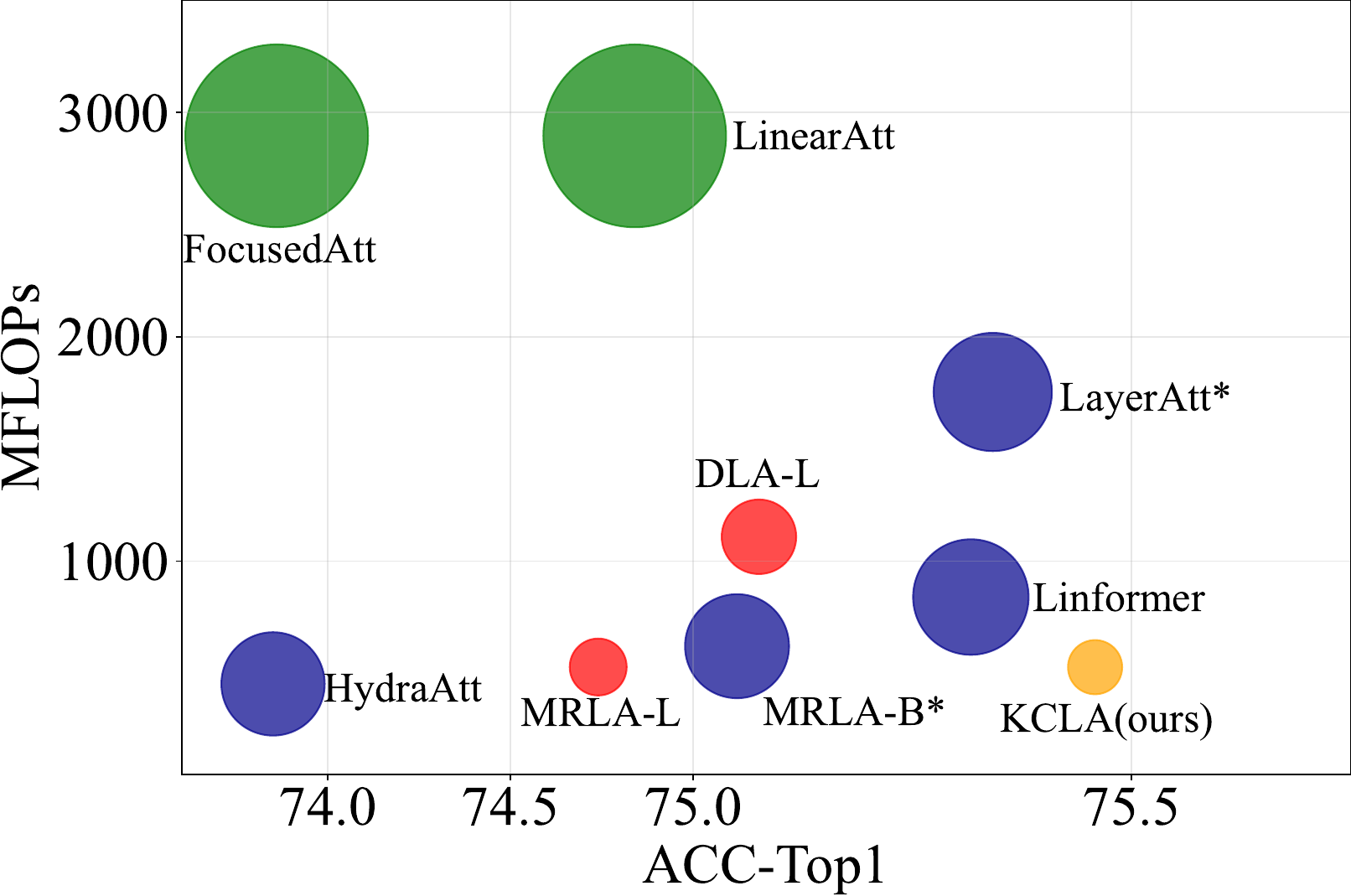}
    \caption{Comparison of different layer attention methods. * marks methods with $\mathcal{O}(D^2)$ complexity. A blue circle indicates methods requiring caching previous layer outputs, and a green circle indicates methods requiring calculation and storage of state $\boldsymbol{K}^\top\boldsymbol{V}$. The size of circles represents memory allocation.}
    \label{fig:comparision-with-other-linearAttention}
\end{figure}

To mitigate the limitation of previous methods, we propose Key-Correlated Layer Attention (KCLA), a novel linearized layer attention method. KCLA is motivated by the empirical observation shown in Fig.~\ref{fig:layer-att-ksim-summary} that Key representations in standard layer attention across different layers exhibit strong correlations. Building on this insight, KCLA linearizes the standard layer attention by linearly approximating Key representations, avoiding the problem of static representations in RLA. The linear approximation not only retains the ability to establish long-range dependencies, but also enables a novel approximation of the indispensable softmax operation used in the standard attention mechanism. In Fig.~\ref{fig:comparision-with-other-linearAttention}, we compared the performance of KCLA and other methods on the CIFAR-100 dataset, including several techniques originally designed for standard attention that were adapted and implemented here in the form of layer attention. KCLA achieved the best balance among accuracy, memory cost, and computational complexity. The main contributions of this work can be summarized as follows:
\begin{itemize}
    \item We observe and prove that the directions of Key representations across different layers in standard layer attention are highly correlated.
    \item We propose KCLA, a novel linearized layer attention method that addresses the challenges of linearizing the standard layer attention. 
    \item For KCLA, we propose a novel softmax approximation method, achieving dynamic and adaptive adjustment of the attention distribution at each layer.
\end{itemize}

\section{Related work}
\subsection{Cross-layer interaction}
Skip connections underscore the critical role of cross-layer information interaction in various tasks \cite{7780459,niu2024bidirectional,xu2026cnm,yu2026mobileode,he2024lightweight,cao2026enhancing,he2020subtraction,he2023cascade,2015U}. While fixed-position skip connections effectively facilitate gradient propagation across deep layers \cite{whyisskipconnection}, offering limited flexibility, more advanced methods propose flexible cross-layer feature aggregation. Recurrent neural networks, such as RLANet \cite{zhao2021recurrence}, borrowed concepts from time series analysis for lightweight layer aggregation. Separately, methods employing layer attention have emerged. RLA \cite{fang2023crosslayerretrospectiveretrievinglayer} introduced layer attention by treating DNN layers as sequence tokens. Addressing RLA's static nature, DLA \cite{wang2024strengthening} proposed a dynamic sharing unit. Beyond these, other flexible methods utilize different models: DIANet \cite{huang2019dianetdenseandimplicitattentionnetwork} incorporated LSTM for inter-layer relationships, and GLA \cite{10743821} further advanced with a layer attention mechanism for efficiency. State space models have also been applied for inter-layer aggregation in S6LA \cite{liu2025layers}. Additionally, some work focuses on modifying standard residual connections, such as introducing weighted factors \cite{zhang2024residual} to adjust identity shortcut impact. \cite{xiao2025muddformer} introduced the MUDDFormer model, which utilizes Multiway Dynamic Dense (MUDD) to enhance cross-layer information flow and address residual connection limitations. \cite{li2025enhancing} improved layer attention efficiency and performance by quantifying and removing its redundancy.

\subsection{Linear Attention}
Standard softmax attention suffers from quadratic $O(N^2)$ complexity with respect to the sequence length $N$. This limitation has driven extensive research into efficient linear alternatives. A foundational method involves linear attention \cite{katharopoulos2020transformersrnnsfastautoregressive}, which approximates the similarity matrix using a kernel $\text{softmax}(\boldsymbol{Q}\boldsymbol{K}^\top)\approx\phi(\boldsymbol{Q})\phi(\boldsymbol{K})^\top$. By reordering the computation via matrix associativity, $\text{Attention}(\boldsymbol{Q}, \boldsymbol{K}, \boldsymbol{V}) = \phi(\boldsymbol{Q})\left(\phi(\boldsymbol{K})^\top \boldsymbol{V}\right)$, this approach achieves $O(N)$ complexity. Methods following this paradigm, such as linear attention~\cite{katharopoulos2020transformersrnnsfastautoregressive}, Gated Linear Attention~\cite{yang2024gatedlinearattentiontransformers}, and FLatten Transformer~\cite{han2023flatten}, linearize attention by reordering computations and maintaining a fixed-size intermediate state $(\boldsymbol{K}^\top \boldsymbol{V}) \in \mathbb{R}^{d \times d_v}$. However, applying this method, particularly to layer attention, can result in high memory consumption and slow inference due to the typically large dimensions of $d$ and $d_v$. Compounding this issue, the graph tensor introduces additional dimensions $H$ and $W$, rendering the memory consumption of the intermediate product $\boldsymbol{K}^\top \boldsymbol{V}$ even more prohibitive. Beyond the kernel approximation paradigm ($\phi(\boldsymbol{Q})\phi(\boldsymbol{K})^\top$), other techniques reduce attention complexity through different approximation or sparsification strategies. Examples include low-rank approximation of $\boldsymbol{K}$ and $\boldsymbol{V}$, a technique used in Linformer \cite{wang2020linformer}; locality-sensitive hashing, as utilized by Reformer \cite{kitaev2020reformer}; and Vector Quantization on $\boldsymbol{K}$, which is employed in Transformer-VQ \cite{lingle2023transformer}.


\section{Problem formulation}
\subsection{Review standard layer attention and recurrent layer attention}
Layer attention \cite{fang2023crosslayerretrospectiveretrievinglayer} is defined as using the Query representation of the $t$-th layer to query information from previous layers to achieve hierarchical interaction. Let $\boldsymbol{X}^{t} \in \mathbb{R}^{1\times d}$ denote the output of $t$-th layer of the network. The outputs of the first $t$ layers form a sequence, i.e. $\left(\boldsymbol{X}^{1}, \ldots,\boldsymbol{X}^{t}\right)$. The $\boldsymbol{Q}^t\in \mathbb{R}^{1\times d}$, $\boldsymbol{K}^t \in \mathbb{R}^{t\times d}$ and $\boldsymbol{V}^t \in \mathbb{R}^{t\times d_v}$ are calculated as follows:
\begin{equation}
\left\{
    \begin{aligned}
    \boldsymbol{Q}^t&=f^t_{Q}(\boldsymbol{X}^t)
    \\\boldsymbol{K}^t&=f_{K}^t(\text{Concat}\left[\boldsymbol{X}^1, \ldots, \boldsymbol{X}^t\right])
    \\
    \boldsymbol{V}^t&=f_{V}^t(\text{Concat}\left[\boldsymbol{X}^1, \ldots, \boldsymbol{X}^t\right])
    \end{aligned}
    \label{eq:basic method to compute K and V}
\right.
\end{equation}
Let $f_Q^t$, $f_K^t$, and $f_V^t$ be the learnable linear transformations at the $t$-th layer, mapping $\mathbb{R}^d \to \mathbb{R}^d$, $\mathbb{R}^d \to \mathbb{R}^d$, and $\mathbb{R}^d \to \mathbb{R}^{d_v}$ respectively. The attention output $\boldsymbol{O}^t\in\mathbb{R}^{1\times d_v}$ for the $t$-th layer in the network is calculated as follows:
\begin{equation}
    \boldsymbol{O}^t = \mathrm{softmax}\left(\frac{\boldsymbol{Q}^t(\boldsymbol{K}^t)^{\top}}{\sqrt{d}}\right)\boldsymbol{V}^t
    \label{eq:LA output formulation}
\end{equation}
To reduce repeated calculations, RLA~\cite{fang2023crosslayerretrospectiveretrievinglayer} only applies $f^t_{K}(\cdot)$ and $f^t_{V}(\cdot)$ to the output of the current layer $\boldsymbol{X}^t$, and then concatenate them with the cached $\boldsymbol{K}^{t - 1}$ and $\boldsymbol{V}^{t - 1}$ as follows:
{
\begin{equation}
\left\{
\begin{aligned}
\boldsymbol{K}^t&=\text{Concat}\left[\boldsymbol{K}^{t-1}, f_{K}^t(\boldsymbol{X}^t)\right]\\
\boldsymbol{V}^t&=\text{Concat}\left[\boldsymbol{V}^{t-1} f_{V}^t(\boldsymbol{X}^t)\right]
\end{aligned}
\label{eq:RLA's calculate KV}
\right.
\end{equation}
}
Since the Key representation and Value representation of each layer remain unchanged after caching, RLA directly mixes the low-level high-frequency features of the early layers with the subsequent layers, impairing the late-stage feature abstraction ability \cite{zhang2024residual}.

\subsection{Limitation of MRLA-L}
\begin{figure}
    \centering 
    \subfloat[]{%
        \includegraphics[width=0.49\linewidth]{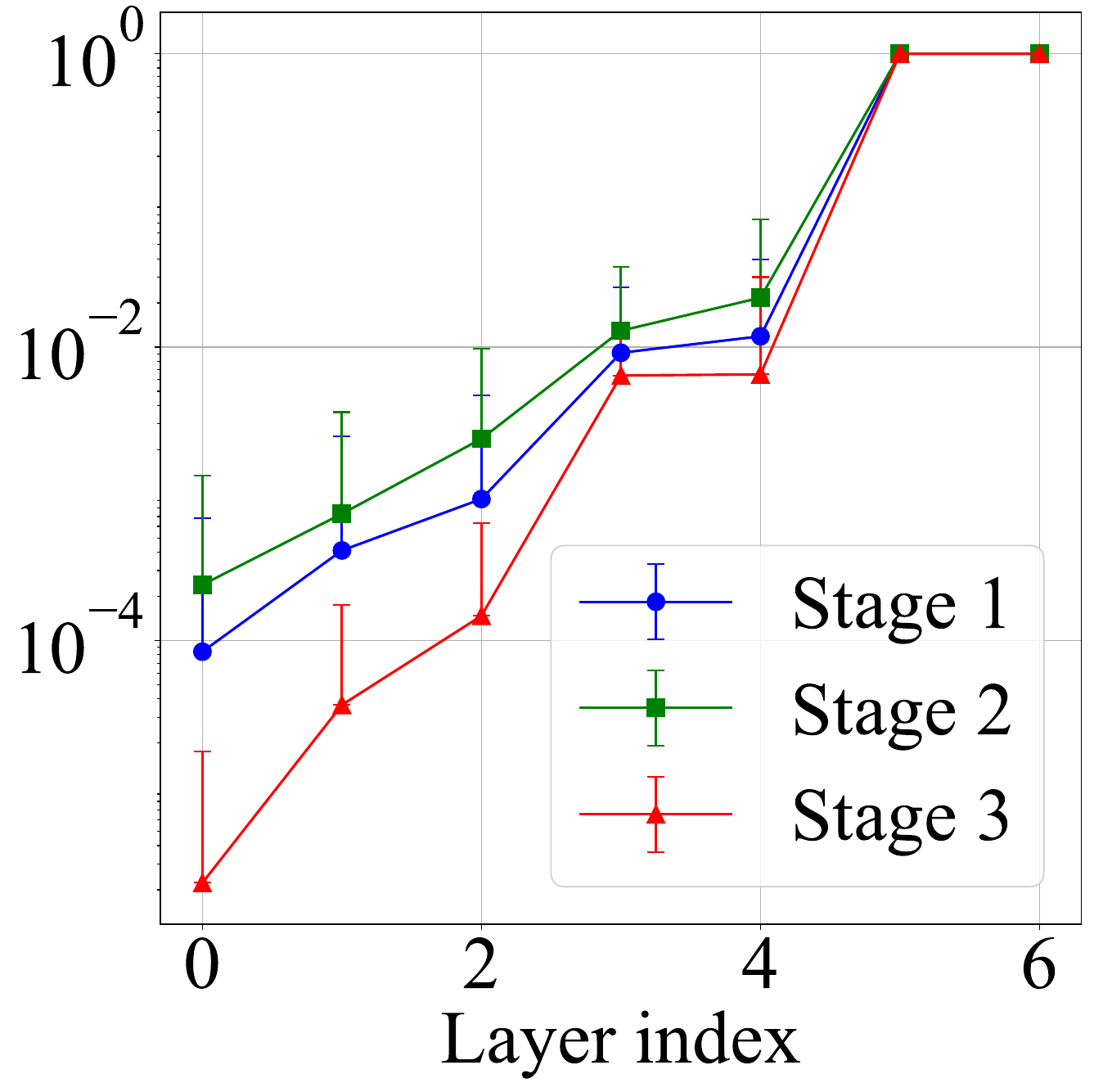}%
    }
    \subfloat[]{%
        \includegraphics[width=0.49\linewidth]{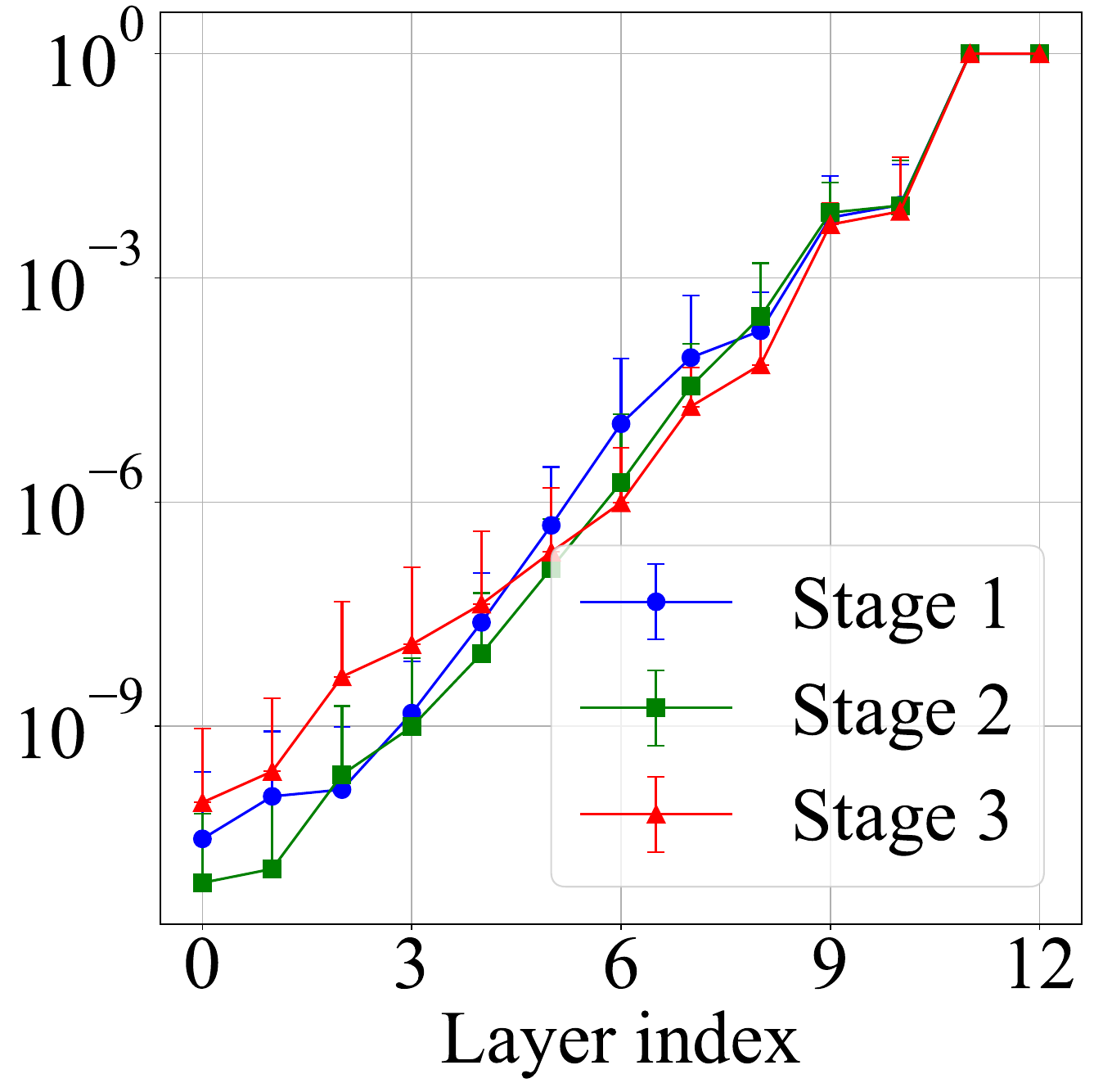}%
    }\\
    \caption{The means and standard deviations of $\boldsymbol{\beta}$ for different layers in MRLA-L, with error bars representing the corresponding standard deviations. The lower part of the error bars is truncated where it would exceed the figure boundary. Among them, (a) is tested at a depth of 56, and (b) is tested at a depth of 110.}
    \label{fig: beta in MRLA-L}
\end{figure}
    
MRLA-L~\cite{fang2023crosslayerretrospectiveretrievinglayer} is a lightweight version of RLA, which omits the softmax normalization operation for attention scores, and divides the attention output of RLA into two parts: the query of the current layer to the previous $t-1$ layers  and the query of the current layer to itself, denote as  $\boldsymbol{Q}^t(\boldsymbol{K}^t_{[1,t-1],:})^\top\boldsymbol{V}^t_{[1,t-1],:}$ and $\boldsymbol{Q}^t(\boldsymbol{K}^t_{t,:})^\top\boldsymbol{V}^t_{t,:}$ respectively. Here $\boldsymbol{K}^t_{[1,t - 1],:}$ represents the Key representation of the first $t - 1$ layers, and $\boldsymbol{K}^t_{t,:}$ represents the Key representation of the $t$-th layer. Due to the static structure of RLA, i.e., $\boldsymbol{K}_{i,:}^{t - 1}=\boldsymbol{K}^t_{i,:}$ and $\boldsymbol{V}_{i,:}^{t - 1}=\boldsymbol{V}^t_{i,:}$ for $i\in[1,t - 1]$, it can be deduced that $\boldsymbol{Q}^t(\boldsymbol{K}^t_{[1,t-1],:})^\top\boldsymbol{V}^t_{[1,t-1],:}=\boldsymbol{Q}^t\boldsymbol{K}^{t - 1}\boldsymbol{V}^{t - 1}$. MRLA-L reduces the computational complexity by linearly approximating $\boldsymbol{Q}^t\boldsymbol{K}^{t - 1}\boldsymbol{V}^{t - 1}$ as follows:
\begin{equation}
    \begin{aligned}
    \boldsymbol{Q}^{t} (\boldsymbol{K}^{t - 1})^\top \boldsymbol{V}^{t - 1} &= (\boldsymbol{\lambda}_q^{t} \odot \boldsymbol{Q}^{t - 1})(\boldsymbol{K}^{t - 1})^\top \boldsymbol{V}^{t - 1} 
    \\
    &\approx \boldsymbol{\lambda}_o^{t} \odot [\boldsymbol{Q}^{t - 1} (\boldsymbol{K}^{t - 1})^\top \boldsymbol{V}^{t - 1}]
    \\
    &= \boldsymbol{\lambda}_o^{t} \odot \boldsymbol{O}^{t - 1}
\end{aligned}
\end{equation}
Here $\odot$ denotes the Hadamard product. $ \boldsymbol{\lambda}_q^t\in\mathbb{R}^{1\times d}$ and  $\boldsymbol{\lambda}_o^t\in \mathbb{R}^{1\times d_v}$ respectively. The validity of this approximation is based on the observation in the experiment that the Query representations of adjacent layers in RLA have a similar pattern. When $\boldsymbol{Q}^t$ is completely correlated with $\boldsymbol{Q}^{t - 1}$, all elements in $\boldsymbol{\lambda}_o^{t}$ are equal, and the approximation becomes an equality. The attention output of MRLA-L is as follows:
\begin{equation}
\begin{aligned}
    \boldsymbol{O}^t & = \boldsymbol{\lambda}_o^t \odot \boldsymbol{O}^{t - 1} + \boldsymbol{Q}^t (\boldsymbol{K}_{t,:}^t)^\top \boldsymbol{V}_{t,:}^t
\\
&= \sum_{l = 0}^{t - 1} \boldsymbol{\beta}_l \odot \big[ \boldsymbol{Q}^{t - l} (\boldsymbol{K}_{t - l,:}^{t - l})^\top \boldsymbol{V}_{t - l,:}^{t - l} \big]
\end{aligned}
\label{eq:mrla-l-beta-sum}
\end{equation}
where $\boldsymbol{\beta}_0 = \mathbf{1}$, and $\boldsymbol{\beta}_l = \boldsymbol{\lambda}_o^t \odot \dots \odot \boldsymbol{\lambda}_o^{t - l + 1}$ for $l \geq 1$, $\boldsymbol{\lambda}_o^t$ is specified as learnable parameters. 
We trained MRLA-L using ResNet-50 and ResNet-110 as the backbones respectively to observe $\boldsymbol{\lambda}_o$ at different layers, and calculated $\boldsymbol{\beta}_{l}$ and the corresponding mean and standard deviation for each layer. The results are shown in Fig.~\ref{fig: beta in MRLA-L}. When calculating the layer attention output of the final layer in each stage, $\boldsymbol{\beta}$ of the early layers within the same stage have become very small. Consequently, the contributions of these early layers in Equation~(\ref{eq:mrla-l-beta-sum}) are almost negligible, which demonstrates MRLA-L's inability to establish effective long-range cross-layer connections.

\begin{figure*}[ht]
    \centering 
    \includegraphics[width=0.89\textwidth]{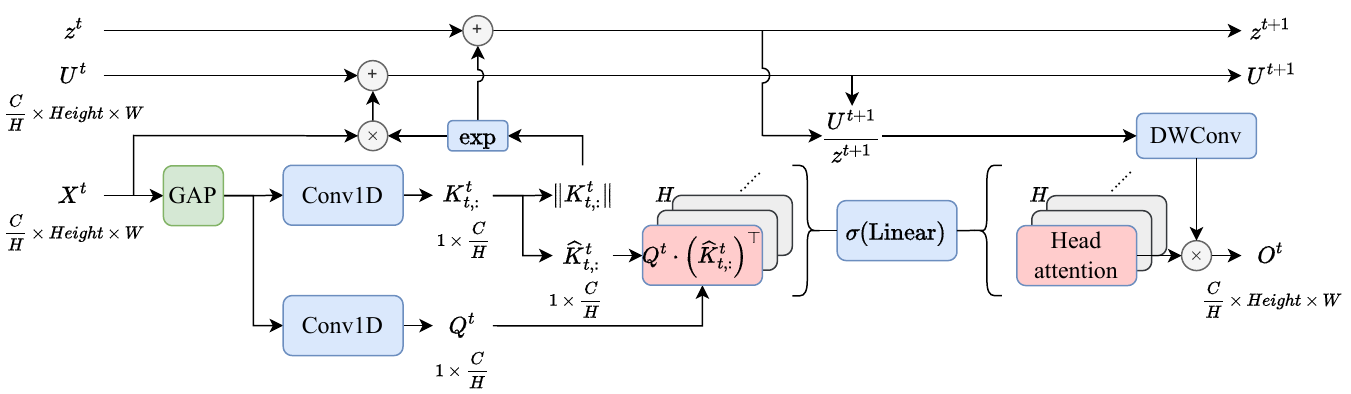}%
    \caption{Implementation details of KCLA.}
    \label{fig:kcla_attention_details}
\end{figure*}

\section{Proposed Key-Correlated layer attention}\label{sec:Proposed Method}
Our analysis of layer attention on ResNet revealed consistently strong correlations between Key representations across different layers. Temporarily omitting the softmax operation on attention score, we approximate the formula for the output of layer attention as follows:
{
\begin{equation}
\begin{aligned}
\boldsymbol{O}^t &= \sum_{i=1}^t\boldsymbol{Q}^t(\boldsymbol{K}^t_{i,:})^\top\boldsymbol{V}^t_{i,:}= \|\boldsymbol{Q}^t\|\sum_{i=1}^t\cos\theta_i\|\boldsymbol{K}^t_{i,:}\|f_V^t(\boldsymbol{X}^i)\\
&\approx \|\boldsymbol{Q}^t\|\cos\theta\sum_{i=1}^{t}\|\boldsymbol{K}^t_{i,:}\|f^t_V(\boldsymbol{X}^i)\\
&= \|\boldsymbol{Q}^t\|\cos\theta\, f^t_V\left(\sum_{i=1}^{t}\|\boldsymbol{K}^t_{i,:}\|\boldsymbol{X}^i\right)
\end{aligned}
\label{eq:KCLA output}
\end{equation}
}
Since $f^t_V(\cdot)$ is a linear transformation acting on all layer outputs. We factor $f^t_V(\cdot)$ out of the summation term. This is mathematically equivalent while retaining the dynamics of the standard layer attention, as the early layer outputs will also be transformed. When the Key representations of different layers are completely correlated, the approximation in Equation~(\ref{eq:KCLA output}) becomes an equality. Therefore, we employ the standard multi-head attention (MHA) mechanism. MHA divides the inputs into $H$ heads, computes attention independently for each head, and concatenates the resulting outputs. A consequence of using MHA is that when the number of heads $H$ equals the input dimension $d$, the dimension of a single head becomes $d/H=1$. In this specific case, the Key representation can be considered completely correlated (in the absence of zero elements). More generally, even when the dimension of a single head is less than $d$ (i.e., $H>1$), the reduced dimensionality makes the linear approximation assumption easier to satisfy. Let $\boldsymbol{U}^t=\sum_{i = 1}^{t}\|\boldsymbol{K}^t_{i,:}\|\boldsymbol{X}^i$. By adopting a recurrent form for $\|\boldsymbol{K}^t_{i,:}\|$, i.e., $\|\boldsymbol{K}^{t-1}_{i,:}\|=\|\boldsymbol{K}^t_{i,:}\|$, we can write Equation~(\ref{eq:KCLA output}) in an iterative form:
\begin{equation}
\begin{aligned}
\boldsymbol{O}^t = \|\boldsymbol{Q}^t\|\cos\theta f^t_V\left(\boldsymbol{U}^t\right),
\boldsymbol{U}^t = \boldsymbol{U}^{t-1} + \|\boldsymbol{K}^t_{t,:}\|\boldsymbol{X}^t
\end{aligned}
\label{eq:iterative update U}
\end{equation}

\subsection{Restore the softmax}
Softmax provides essential attention score normalization in attention mechanisms \cite{46201}. While omitted in Section~\ref{sec:Proposed Method}, applying softmax to KCLA leads to:
\begin{equation}
\left\{
\begin{aligned}
\text{softmax}\left(\|\boldsymbol{Q}^t\|\cos\theta\cdot\left[\|\boldsymbol{K}^t_{1,:}\|,\ldots,\|\boldsymbol{K}^t_{t,:}\|\right]\right)_j \qquad
\\ 
= \frac{\exp\left(\|\boldsymbol{Q}^t\|\cos\theta \|\boldsymbol{K}^t_{j,:}\|\right)}{\sum_{i=1}^t \exp\left(\|\boldsymbol{Q}^t\|\cos\theta \|\boldsymbol{K}^t_{i,:}\|\right)}
\end{aligned}
\label{eq:softmax_on_KCLA}
\right.
\end{equation}
Note that the angles of Key representations are approximated as equal in KCLA. 
In Equation~(\ref{eq:softmax_on_KCLA}),  $\|\boldsymbol{Q}^t\|\cos\theta\in\mathbb{R}$ serves as the common coefficient for all terms, playing a role similar to a temperature parameter \cite{NEURIPS2024_d2fe3a57,oorloff2025mitigating}. Larger values result in a sharper distribution, while smaller values yield a smoother distribution. And $\|\boldsymbol{K}^t_{j,:}\|\in\mathbb{R}$ describes the importance of the output of the $j$-th layer. Equation~(\ref{eq:softmax_on_KCLA}) requires independently saving the norms of the outputs of each previous layer. However, in Equation~(\ref{eq:iterative update U}), $\|\boldsymbol{K}^t_{j,:}\|$ has already been accumulated into $\boldsymbol{U}^t$. Although we can apply the exponential transformation to $\|\boldsymbol{K}^t_{t,:}\|$ when updating $\boldsymbol{U}^t$, KCLA still loses the ability to dynamically adjust the attention score distribution according to $\|\boldsymbol{Q}^t\|\cos\theta$.

To address this issue, we propose a modified multi-head attention mechanism. By setting different scaling factors for different heads, we obtain heads with different temperatures. Then, we use $\|\boldsymbol{Q}^t\|\cos\theta$ to weight each head to achieve dynamic temperature adjustment. Specifically, since scaling factors are parameters with multiplicative effects, whose behavior better aligns with changes on a logarithmic scale \cite{bergstra2012random}, we first generate a logarithmically uniform sequence $\mathcal{P}$ in the range $(p,1/p)$ with a length equal to $H$, where $H$ is the number of heads and $p\in(0,1)$. Then, replace the original fixed scaling factor $\sqrt{d/H}$ for each head with $\mathcal{D} = \mathcal{P}\cdot\sqrt{d/H}$. After introducing the exponential transformation and heads with different temperatures, $\boldsymbol{U}^t$ in Equation~(\ref{eq:iterative update U}) and the normalization term $\boldsymbol{z}^t$ in head $h$ are formulated as follows:
{
\begin{equation}
\left\{
\begin{aligned}
\boldsymbol{U}^t_h &= \sum_{i=1}^{t} \exp\left(\frac{\|\boldsymbol{K}^t_{h[i,:]}\|}{\mathcal{D}_h}\right) \boldsymbol{X}^i_h\\
&= \boldsymbol{U}^{t-1}_h +\exp\left(\frac{\|\boldsymbol{K}^t_{h[t,:]}\|}{\mathcal{D}_h}\right)\boldsymbol{X}^t_h\\
\boldsymbol{z}^t_h &= \sum_{i=1}^{t} \exp\left(\frac{\|\boldsymbol{K}^t_{h[i,:]}\|}{\mathcal{D}_h}\right)\\
&= \boldsymbol{z}^{t-1}_h + \exp\left(\frac{\|\boldsymbol{K}^t_{h[t,:]}\|}{\mathcal{D}_h}\right)
\end{aligned}
\right.
\label{eq:iterative update U with softmax}
\end{equation}
}
To restore the ability of $\|\boldsymbol{Q}^t\|\cos\theta$ to dynamically adjust the attention distribution, we feed $\|\boldsymbol{Q}^t\|\cos\theta$ into a learnable linear transformation $\Phi: \mathbb{R}^{H} \to \mathbb{R}^{H}$. The output $\Phi(\|\boldsymbol{Q}^t\|\cos\theta)$ serves as the head attention score for the heads with different temperatures, thus indirectly enabling $\|\boldsymbol{Q}^t\|\cos\theta$ to dynamically regulate the attention distribution. The attention output for head $h$ is:
\begin{equation}
\boldsymbol{O}^t_h = \underbrace{\text{sigmoid}\left(\Phi\left(\|\boldsymbol{Q}^t\|\cos\theta\right)\right)_h}_{\text{Head attention}} \cdot \underbrace{f^t_V\left(\frac{\boldsymbol{U}^t_h}{\boldsymbol{z}^t_h}\right)}_{\text{Normalized Value}}
\label{eq:KCLA_multihead_output}
\end{equation}

Fig.~\ref{fig:kcla_attention_details} illustrates the details of KCLA in CNNs. It processes the backbone layer output $\boldsymbol{X}^t$ using global average pooling (GAP). Linear transformations $f^t_Q(\cdot)$, $f^t_K(\cdot)$, $f^t_V(\cdot)$ are implemented using 1D convolution (Conv1D) and $3\times3$ depth-wise convolution (DWConv) to derive Query, Key, and normalized Value representations. The linear transformation $\Phi$ is implemented using a linear layer to transform the per-head dot products $\boldsymbol{Q}^t(\hat{\boldsymbol{K}}^t_{t,:})^\top$ to obtain head attention scores. $\hat{\boldsymbol{K}}^t_{t,:}$, $H$, and $\sigma$ denote the unit direction vector of $\boldsymbol{K}^t_{t,:}$, the number of heads, and the sigmoid activation function, respectively.

\subsection{The cross stage design}
\label{sec:cross stage degisn}
Some neural networks are structured into multiple stages, where varying numbers of channels and feature map sizes are employed. This structural characteristic poses challenges for the application of layer attention, as the shape of each token within the attention mechanism must remain consistent. In KCLA, when transitioning between stages, the features from the preceding stage have already been aggregated into $\boldsymbol{U}^t$, and the normalization factors accumulated in $\boldsymbol{z}^t$. To bridge the dimensional disparity that arises from the differing feature map sizes and channel counts of the subsequent stage, both $\boldsymbol{U}^t$ and $\boldsymbol{z}^t$ undergo adaptation through a dedicated procedure. The specific procedure for this cross-stage adaptation is detailed in Algorithm~\ref{alg:cross_stage_design}.

\begin{algorithm}
\caption{Adjust $\boldsymbol{U}^t$ and $z^t$ when crossing stages}
\begin{algorithmic}[1]

  \Statex \textbf{Input:} Aggregated features $\boldsymbol{U}^t \in \mathbb{R}^{B\times g\times h\times H\times W}$;
  \Statex \hspace{1.5em} Per-head normalization factors $\boldsymbol{z}^t \in \mathbb{R}^{B\times g}$;
  \Statex \hspace{1.5em} $1\times1$ conv $Conv_{1\times1}$ ($h\!\to\!C_{int}$, stride=2);
  \Statex \hspace{1.5em} Adaptive pooling $AP_S$ ($1\times1$ output);
  \Statex \hspace{1.5em} Transform $f_k:\mathbb{R}^{gC_{int}}\to\mathbb{R}^{gh}$.

  \Statex \textbf{Output:} Processed $\boldsymbol{U}^1 \in \mathbb{R}^{B\times g\times h\times H\times W}$; normalization $\boldsymbol{z}^1_{exp} \in \mathbb{R}^{B\times g\times1}$.

  \Statex \Comment{--- Normalize accumulated features per head ('g') ---}
  \State $\boldsymbol{z}^t_{exp} \gets \text{Unsqueeze}(\boldsymbol{z}^t, (2,3,4))$
  \State $\boldsymbol{U}_{norm} \gets \boldsymbol{U}^t / \boldsymbol{z}^t_{exp}$

  \Statex \Comment{--- Stage transition processing ---}
  \State $\boldsymbol{X}_{in\_q} \gets \text{Reshape}(\boldsymbol{U}_{norm}, (B\!\cdot\!g, h, H, W))$
  \State $\boldsymbol{X}_{conv} \gets Conv_{1\times1}(\boldsymbol{X}_{in\_q})$
  \State $\boldsymbol{q} \gets AP_S(\boldsymbol{X}_{conv})$

  \Statex \Comment{--- Compute new per-head normalization $\boldsymbol{z}^1_{exp}$ ---}
  \State $\boldsymbol{K}_{repr} \gets f_k(\boldsymbol{q})$
  \State $\boldsymbol{K} \gets \text{Reshape}(\boldsymbol{K}_{repr}, (B,g,h))$
  \State $\boldsymbol{k}_{norm} \gets \|\boldsymbol{K}\|_{L2,\text{dim}=2}$
  \State $\boldsymbol{z}^1 \gets \boldsymbol{k}_{norm}$
  \State $\boldsymbol{z}^1_{exp} \gets \exp(\boldsymbol{z}^1)$

  \Statex \Comment{--- Apply new weights ---}
  \State $\boldsymbol{z}^1_{exp} \gets \text{Unsqueeze}(\boldsymbol{z}^1_{exp}, (2,3,4))$
  \State $\boldsymbol{U}^1 \gets \boldsymbol{z}^1_{exp} \odot \boldsymbol{U}_{norm}$
\end{algorithmic}
\label{alg:cross_stage_design}
\end{algorithm}

\begin{table}[ht]
\caption{Results on CIFAR-10 and CIFAR-100. \textbf{Bold} and \underline{underlined} indicate the best and the second best performance, respectively.}
\label{tab:Result on CIFAR-100 and CIFAR-10}
\begin{tabularx}{\linewidth}{lXXXX}
\toprule
\multirow{2}{*}{Model}& \multicolumn{2}{c}{CIFAR-10}&\multicolumn{2}{c}{CIFAR-100}\\
\cmidrule(lr){2-3} \cmidrule(lr){4-5}
&Params(M) & Top1(\%)&Params(M) & Top1(\%)\\
\midrule
\textbf{ResNet-20} &0.22& $91.46$& 0.24 & $69.22$ \\
+SE &0.24&$91.82$& 0.27 &$70.60$\\
+DIANet &0.44&$92.14$& 0.46&$70.01$\\
+Layer Att &0.23&$92.96$& 0.25 & $\underline{72.09}$\\
+MRLA-L &0.23&$92.89$& 0.25 & $71.36$\\
+MRLA-B &0.23&$\underline{93.05}$& 0.25 & $72.05$\\
+DLA-L &0.41&$92.95$& 0.43 & $71.84$\\
+KCLA(ours) &0.23&$\mathbf{93.06}$& 0.26 &$\mathbf{72.25}$\\
\midrule
\textbf{ResNet-56}&0.59&$93.84$& 0.61 & $73.27$ \\
+SE &0.66&$94.09$& 0.68 &$74.58$\\
+DIANet &0.81&$94.32$& 0.83&$74.69$\\
+Layer Att &0.62&$94.63$& 0.64 & $\underline{75.31}$\\
+MRLA-L &0.62&$94.55$& 0.65 & $74.74$\\
+MRLA-B &0.62&$94.45$& 0.64 & $75.12$\\
+DLA-L &0.80&$\mathbf{94.72}$& 0.82 & $ 75.18$\\
+KCLA(ours) &0.63 &$\underline{94.64}$& 0.65 & $\mathbf{75.45}$\\
\midrule
\textbf{ResNet-110} &1.17&$94.45$& 1.17 & $75.34$\\
+SE &1.28&$94.74$& 1.30 &$76.35$\\
+DIANet &1.37&$\underline{94.98}$& 1.39&$76.54$\\
+MRLA-L &1.21&$94.91$& 1.24 & $75.81$\\
+DLA-L &1.39&$\mathbf{95.12}$& 1.41 & $\underline{76.82}$\\
+KCLA(ours) &1.22&$ 94.97$& 1.24 & $\mathbf{76.98}$ \\
\bottomrule
\end{tabularx}
\end{table}

\begin{table}[ht]
\centering
\caption{Results on ImageNet-1K.}
\label{tab:Result_on_ImageNet_1K}
\begin{tabularx}{\linewidth}{lXXXX}
\toprule
Model &Params(M) & FLOPs(G) & Top1(\%) & Top5(\%)\\
\midrule
\textbf{ResNet-50}& 25.6  & 4.1  & 76.1 & 92.9\\
+SE & 28.1  & 4.1  & 76.7 & 93.4\\
+CBAM & 28.1  & 4.2  & 77.3 & 93.7\\
+AA& 27.1  & 4.5  & 77.7 & $\underline{93.8}$\\
+ECA & 25.6  & 4.1  & 77.5 & 93.7\\
+DIANet& 28.4  & - & 77.2 & -\\
+RLA$_g$ & 25.9 & 4.5  & 77.2 & 93.4\\
+MRLA-L & 25.7  & 4.2  & 77.7 & $\underline{93.8}$\\
+DLA-L&27.2 & 4.3 & $\mathbf{78.0}$ & $\mathbf{94.0}$\\
+KCLA(ours)&25.8&4.2&$\underline{77.8}$&$\underline{93.8}$ \\ 
\midrule
\textbf{ResNet-101} & 44.5  & 7.8  & 77.4 & 93.5\\
+SE & 49.3  & 7.8  & 77.6 & 93.9\\
+CBAM & 49.3 & 7.9  & 78.5 & 94.3\\
+AA & 47.6  & 8.6  & 78.7 & $\underline{94.4}$\\
+ECA & 44.5  & 7.8  & $78.7$ & 94.3 \\
+RLA$_g$ & 45.0  & 8.4  & 78.5 & 94.2 \\
+MRLA-L & 44.9  & 7.9  & 78.7 & $\underline{94.4}$ \\
+DLA-L & 47.8  & 8.1  & $\mathbf{78.9}$ & $\mathbf{94.5}$\\
+KCLA(ours)&45.0&8.0&$\underline{78.8}$&$\underline{94.4}$\\ 
\bottomrule
\end{tabularx}
\end{table}

\section{Experiments}
\label{sec:experiments}
To evaluate the effectiveness of the proposed method, we conducted comprehensive experiments across three fundamental vision tasks: image classification, object detection, and  medical image segmentation. We use ResNet \cite{7780459} as the backbone for the image classification and object detection tasks. For the medical image segmentation task, which is not evaluated by MRLA, we use a UNet with a channel-reduced ResNet-18 encoder as the backbone network. For the experiments conducted on ImageNet-1K \cite{5206848} and COCO datasets, as well as the segmentation experiments, we cite results from existing work to ensure fair comparisons. All seeds in our experiments were set to 42 to ensure reproducibility. The code is publicly available.\footnote{\url{https://github.com/bgx666/KCLA}}

\subsection{Image classification}
\textbf{Dataset.} We conducted tests on the CIFAR-100, CIFAR-10 and ImageNet-1K datasets. CIFAR-100 is a widely used image classification dataset that consists of 100 categories, each containing 600 color images of size 32x32 pixels, totaling 60,000 images. Among these, 50,000 images are designated for training, while 10,000 images are reserved for testing. The CIFAR-10 dataset contains 60,000 32×32 pixel color images in 10 classes, with 50,000 images for the training set and 10,000 images for the test set. It covers various categories such as airplanes, cars, etc., and is often used for the training and evaluation of image classification algorithms. ImageNet-1K comprises 1,000 categories, with approximately 1.2 million training images, 50,000 validation images, and 100,000 test images. 

\textbf{Settings.}  For the image classification benchmark, we employed three widely-used datasets: CIFAR-10, CIFAR-100 \cite {Krizhevsky2009LearningML} and ImageNet-1K. The experimental augmentation methods and hyperparameters mainly follow the settings in ResNet \cite{7780459}. For CIFAR-100 and CIFAR-10, we utilized standard data augmentation techniques, including random horizontal flipping, padding each side of the image by 4 pixels, and randomly cropping the image to a size of 32×32. Additionally, normalization was applied based on the dataset's mean and standard deviation. The batch size is set to 128, and the initial learning rate is set to 0.1. A MultiStepLR schedule is employed, with the model trained for a total of 180 epochs. In the first ten epochs, linear learning rate warm-up is adopted, and the learning rate linearly increases from 0.001 to 0.1. The learning rate is reduced at the 100th and 150th epochs. The optimizer used is Stochastic Gradient Descent (SGD) with a weight decay of 1e-4 and a momentum of 0.9. All experiments were repeated five times under the same settings, and the mean were reported.

For ImageNet-1K, we followed the data augmentation strategies described in \cite{7780459}  and \cite{fang2023crosslayerretrospectiveretrievinglayer}. During training, images were randomly cropped to 224×224 pixels and horizontally flipped. During evaluation, images were first resized to 256×256 pixels and then center-cropped to 224×224 pixels. The training was conducted for 100 epochs with a batch size of 256. A warm-up strategy was employed for the first $3$ epochs, during which the learning rate was linearly increased from $1\mathrm{e}{-4}$ to the base rate of $0.1$. This was followed by the MultiStepLR schedule, the learning rate was reduced by a multiplicative factor $\gamma$ of $0.1$. Optimization was performed using SGD with a momentum of $0.9$ and a weight decay of $1\mathrm{e}{-4}$. The dimension of each head is set to 16 on the CIFAR-10 and CIFAR-100 dataset, and 32 on the ImageNet-1K dataset, in order to keep the number of parameters as close as possible to that of MRLA. 

\begin{table*}[ht]
\caption{Object detection results of different methods on COCO val2017.}
\label{tab:Results on COCO2017}
\begin{tabularx}{\linewidth}{llXXXXXXX}
\toprule
Detector&Model&Params(M)&$AP^{bb}$ (\%)&$AP^{bb}_{50}$ (\%)&$AP^{bb}_{75}$ (\%)&$AP^{bb}_{S}$ (\%)&$AP^{bb}_{M}$ (\%)&$AP^{bb}_{L}$ (\%)\\
\midrule
\multirow{17}{*}{\shortstack{Faster\\ R-CNN}} &\textbf{ResNet-50} \citep{7780459} &  41.5  & 36.4 & 58.2 & 39.2 & 21.8 & 40.0 & 46.2 \\
&+ SE \citep{hu2019squeezeandexcitationnetworks}& 44.0  & 37.7 & 60.1 & 40.9 & 22.9 & 41.9 & 48.2 \\
&+ ECA \citep{wang2020ecanetefficientchannelattention}& 41.5  & 38.0 & 60.6 & 40.9 & 23.4 & 42.1 & 48.0 \\
&+ RLA$_g$ \citep{zhao2021recurrence}& 41.8 & 38.8 & 59.6 & 42.0 & 22.5 & 42.9 & 49.5 \\
&+ BA \citep{Zhang2024BANetBA}& 44.7  & 39.5 & 61.3 & 43.0 & \textbf{24.5} & 43.2 & 50.6 \\
&+ MRLA-B \citep{fang2023crosslayerretrospectiveretrievinglayer}& 41.7  & 40.1 &61.3& 43.8&24.0 &43.9& 52.4\\
&+ MRLA-L \citep{fang2023crosslayerretrospectiveretrievinglayer}& 41.7 & \underline{40.4} & \underline{61.5} & \underline{44.0} & 24.2 & \underline{44.1} & \underline{52.7}\\
&+ DLA-L \citep{wang2024strengthening}& 44.2 & \textbf{40.6} & \textbf{61.6} & \textbf{44.2} & \textbf{24.5} & \textbf{44.2} & \textbf{52.9}\\
&+ KCLA(ours) &42.0  &40.2& 61.4& 43.8& \underline{24.4}& 44.0& 52.2\\
\cmidrule(lr){2-9}
&\textbf{ResNet-101} \citep{7780459}& 60.5  & 38.7 & 60.6 & 41.9 & 22.7 & 43.2 & 50.4 \\
&+ SE \citep{hu2019squeezeandexcitationnetworks}& 65.2 & 39.6 & 62.0 & 43.1 & 23.7 & 44.0 & 51.4 \\
&+ ECA \citep{wang2020ecanetefficientchannelattention}& 60.5 & 40.3 & 62.9 & 44.0 & 24.5 & 44.7 & 51.3 \\
&+ RLA$_g$ \citep{zhao2021recurrence}& 60.9 & 41.2 & 61.8 & 44.9 & 23.7 & 45.7 & 53.8 \\
&+ BA \citep{Zhang2024BANetBA}& 66.4 & 41.7 & 63.4 & 45.1 & 24.9 & 45.8 & 54.0 \\
&+ MRLA-L \citep{fang2023crosslayerretrospectiveretrievinglayer}& 60.9 & \underline{42.0} & \underline{63.1} & 45.7 & 25.0 & 45.8 & \underline{55.4} \\
&+ DLA-L \citep{wang2024strengthening}& 63.4 & \textbf{42.3} & \textbf{63.3} & \underline{45.8} & \underline{25.2} & \underline{46.0} &\textbf{55.5} \\
&+ KCLA(ours)&61.2&\underline{42.0} &\textbf{63.3}& \textbf{46.0} &\textbf{25.3} &\textbf{46.4} &54.2
\\
\midrule
\multirow{18}{*}{\shortstack{Mask\\R-CNN}}&\textbf{ResNet-50} \citep{7780459}&  44.2 & 37.2 & 58.9 & 40.3 & 22.2 & 40.7 & 48.0 \\
&+ SE \citep{hu2019squeezeandexcitationnetworks}& 46.7& 38.7 & 60.9 & 42.1 & 23.4 & 42.7 & 50.0 \\
&+ ECA \citep{wang2020ecanetefficientchannelattention}& 44.2& 39.0 & 61.3 & 42.1 & 24.2 & 42.8 & 49.9 \\
&+ 1 NL \citep{8578911}& 46.5  & 38.0 & 59.8 & 41.0 & - & - & - \\
&+ GC (r16) \citep{cao2019gcnet}&46.9  & 39.4 & 61.6 & 42.4 & - & - & - \\
&+ GC (r4) \citep{cao2019gcnet}& 54.4& 39.9 & 62.2 & 42.9 & - & - & - \\
&+ RLA$_g$ \citep{zhao2021recurrence}& 44.4 & 39.5 & 60.1 & 43.4 & - & - & - \\
&+ BA \citep{Zhang2024BANetBA}& 47.3& 40.5 & 61.7 & 44.2 & \underline{24.5} & \underline{44.3} & 52.1 \\
&+ MRLA-L \citep{fang2023crosslayerretrospectiveretrievinglayer}&44.3 & \textbf{41.2} & \underline{62.3} & \textbf{45.1} & \textbf{24.8} & \textbf{44.6} & \textbf{53.5} \\
&+KCLA(ours) &44.7&\underline{40.9}& \textbf{62.4}& \underline{44.7} &24.1 &\textbf{44.6} &\underline{53.2}\\
\cmidrule(lr){2-9}
&\textbf{ResNet-101} \citep{7780459}& 63.2  & 39.4 & 60.9 & 43.3 & 23.0 & 43.7 & 51.4 \\
&+ SE \citep{hu2019squeezeandexcitationnetworks}& 67.9 & 40.7 & 62.5 & 44.3 & 23.9 & 45.2 & 52.8 \\
&+ ECA \citep{wang2020ecanetefficientchannelattention}& 63.2 & 41.3 & 63.1 & 44.8 & 25.1 & 45.8 & 52.9 \\
&+ 1 NL \citep{8578911}& 65.5  & 40.8 & 63.1 & 44.5 & - & - & - \\
&+ GC (r16) \citep{cao2019gcnet}& 68.1 & 41.1 & \underline{63.6} & 45.0 & - & - & - \\
&+ GC (r4) \citep{cao2019gcnet}& 82.2 & 41.7 & \textbf{63.7} & 45.5 & - & - & - \\
&+ RLA$_g$ \citep{zhao2021recurrence}& 63.6& 41.8 & 62.3 & 46.2 & - & - & - \\
&+ MRLA-L \citep{fang2023crosslayerretrospectiveretrievinglayer}& 63.5 & \textbf{42.8} & \underline{63.6} & \underline{46.5} & \underline{25.5} & \textbf{46.7} & \underline{55.2} \\
&+ KCLA(ours)&63.9&\underline{42.7}& 63.4& \textbf{46.6}& \textbf{25.6}& \underline{46.5}& \textbf{56.0} \\
\bottomrule
\end{tabularx}
\end{table*}

\textbf{Results on CIFAR-10 and CIFAR-100.}
As shown in Table~\ref{tab:Result on CIFAR-100 and CIFAR-10}, KCLA achieved superior performance over previous methods on CIFAR-100 at all three tested depths: 20, 56, and 110. Furthermore, the parameter count incurred was only marginally larger than the corresponding standard ResNet models, with increases of merely 0.02M, 0.04M, and 0.07M respectively. On the CIFAR-10 dataset, the performance lead of KCLA is slightly diminished. However, it still achieves optimal performance when the network depth is set to 20.

\textbf{Results on ImageNet-1K.}
We conduct a comprehensive comparison between KCLA and the layer attention method MRLA \cite{fang2023crosslayerretrospectiveretrievinglayer}, as well as other attention-based methods, including SE \cite{hu2019squeezeandexcitationnetworks}, CBAM \cite{woo2018cbamconvolutionalblockattention}, AA \cite{9010285}, ECA \cite{wang2020ecanetefficientchannelattention}, DIANet \cite{huang2019dianetdenseandimplicitattentionnetwork}, $RLA_g$ \cite{zhao2021recurrence} and DLA-L \cite{wang2024strengthening}. Experimental results, as presented in Table~\ref{tab:Result_on_ImageNet_1K}, demonstrate that when integrated into ResNet-50 and ResNet-101 architectures, KCLA achieves performance improvements of $+1.7\%$ and $+1.4\%$ in top-1 accuracy respectively over the baseline, outperforming all conventional channel or spatial attention-based modules. Notably, KCLA maintains marginal yet consistent advantages ($+0.1\%$ top-1 accuracy for both network depths) over the linear layer attention architecture MRLA-L, while remaining highly parameter-efficient with only $0.1\text{M}$ additional parameters. Compared to the dynamic layer attention method DLA-L, KCLA's top-1 accuracy is only lower by $0.2\%$ and $0.1\%$ at depths of $50$ and $101$, respectively. However, DLA-L introduces a substantially larger overhead of $1.4\text{M}$ and $2.8\text{M}$ extra parameters. This high parametric cost for a marginal performance edge demonstrates a poor trade-off in DLA-L.

\subsection{Object detection}
\textbf{Dataset.}
The experiment uses the COCO2017 dataset \cite{lin2014microsoft}. COCO2017 is a widely used large-scale computer vision dataset that supports tasks such as object detection, instance segmentation, keypoint detection, and image captioning. The dataset comprises approximately 118,000 training images, 5,000 validation images, and 40,000 test images, covering 80 common object categories.

\textbf{Settings.}
We employed the open-source MMDetection toolkit \cite{chen2019mmdetection} with Faster R-CNN \cite{7485869} and Mask R-CNN \cite{he2017mask} frameworks as detectors. We used Average Precision (AP), a widely adopted evaluation metric for object detection models. The models were initialized with weights pre-trained on ImageNet-1K. The optimization was performed using SGD with a momentum of 0.9 and a weight decay of $1\text{e-}4$. Experiments were conducted on two RTX 4090 GPUs. We adopted the default 1x training strategy schedule, which involves decaying the learning rate by a factor of $\gamma=0.1$ at epochs 8 and 11. Limited by available resources, the batch size was set to 8 for ResNet-50 and 4 for ResNet-101. According to the linear scaling rule, the initial learning rates for these configurations were set to 0.1 and 0.05 respectively. 

\begin{table*}[ht]
\caption{Experimental results of medical image segmentation.}
\label{tab:Results on ISIC2017 and ISIC2018}
\begin{tabularx}{\linewidth}{cXXXXXX}
\toprule
Dataset&Model & Layer attention &Params(M)&FLOPs(G)&mIoU(\%) & DSC(\%) \\
\midrule
\multirow{8}{*}{ISIC2017}
&UNet \cite{2015U} &\qquad$\mathbf{\times}$& 7.77 & 13.76 & 76.98 & 86.99 \\
&UTNetV2 \cite{Gao2022AMT}&\qquad$\mathbf{\times}$ & 12.80 & 15.50 & 77.35 & 87.23 \\
&TransFuse \cite{zhang2021transfuse}&\qquad$\mathbf{\times}$ & 26.16 & 11.50 & 79.21 & 88.40 \\
&UNeXt-S \cite{valanarasu2022unext}&\qquad$\mathbf{\times}$ & 0.32 & 0.10 & 78.26 & 87.80 \\
&MALUNet \cite{ruan2022malunet}&\qquad$\mathbf{\times}$ & 0.18 & 0.09 & 78.78 & 88.13 \\
\cmidrule{2-7}
&ResNet-UNet&\qquad$\mathbf{\times}$& 1.53 & 3.45 & $76.82\pm0.23$ & $86.89\pm0.15$\\
&MRLA-L-UNet&\qquad\checkmark& 1.54 & 3.48 & $78.67\pm0.50$ & $88.06\pm0.31$\\
&DLA-L-UNet&\qquad\checkmark&1.57 & 3.50 & $\boldsymbol{78.96\pm0.31}$ & $\boldsymbol{88.24\pm0.20}$\\
&KCLA-UNet(ours)&\qquad\checkmark& 1.55&3.49 & \underline{$78.71\pm0.32$} & \underline{$88.09\pm0.20$}\\
\midrule
\multirow{10}{*}{ISIC2018}
&UNet \cite{2015U}&\qquad$\mathbf{\times}$& 7.77 & 13.76 & 77.86 & 87.55 \\
&UNet++ \cite{zhou2018unetpp}&\qquad$\mathbf{\times}$& 9.16 & 34.86 & 78.31 & 87.83 \\
&UTNetV2 \cite{Gao2022AMT}&\qquad$\mathbf{\times}$& 12.80 & 15.50 & 78.97 & 88.25 \\
&SANet \cite{wei2021shallow}&\qquad$\mathbf{\times}$& 23.90 & 5.96 & 79.52 & 88.59 \\
&TransFuse \cite{zhang2021transfuse}&\qquad$\mathbf{\times}$& 26.16 & 11.50 & 80.63 & 89.27 \\
&UNeXt-S \cite{valanarasu2022unext}&\qquad$\mathbf{\times}$& 0.32 & 0.10 & 79.09 & 88.33 \\
\cmidrule{2-7}
&ResNet-UNet&\qquad$\mathbf{\times}$& 1.53 & 3.45 & $78.14\pm0.31$ & $87.73\pm0.19$\\
&MRLA-L-UNet&\qquad\checkmark& 1.54 & 3.48 & \underline{$80.36\pm0.20$} & \underline{$89.11\pm0.12$}\\
&DLA-L-UNet&\qquad\checkmark&1.57&3.50&$80.26\pm0.17$ & $89.05\pm0.11$\\
&KCLA-UNet(ours)&\qquad\checkmark& 1.55&3.49 & $\boldsymbol{80.52\pm0.23}$ & $\boldsymbol{89.21\pm0.14}$\\
\midrule
\multirow{4}{*}{Kvasir-SEG}
&UNet \citep{2015U}&\qquad$\mathbf{\times}$ & 13.40&31.12 & $75.52\pm0.78$ & $86.05\pm0.50$\\
&ResNet-UNet&\qquad$\mathbf{\times}$ & 1.53  &3.45 &$72.49\pm0.68$ & $84.05\pm0.45$\\
&MRLA-L-UNet&\qquad\checkmark& 1.54 &3.48 &\underline{$76.59\pm0.80$} & \underline{$86.74\pm0.51$}\\
&DLA-L-UNet&\qquad\checkmark&1.57&3.50&$76.33\pm0.61$ & $86.57\pm0.39$\\
&KCLA-UNet(ours)&\qquad\checkmark& 1.55&3.49 &$\boldsymbol{77.21\pm1.20}$ & $\boldsymbol{87.13\pm0.77}$\\
\bottomrule
\end{tabularx} 
\end{table*}

\textbf{Results on COCO2017.}
In Table~\ref{tab:Results on COCO2017}, when employing ResNet-50 as the backbone, KCLA performs slightly inferior to MRLA-L, which also adopts a layer attention mechanism, leading by only $0.2\%$ in $\text{AP}_{S}^{bb}$. However, this trend is reversed when the depth reaches $101$, with KCLA demonstrating superiority across multiple metrics. Notably, KCLA surpasses the performance of the more complex DLA-L architecture—which introduces an extra module—on $\text{AP}^{bb}_{75}$, $\text{AP}^{bb}_{S}$, and $\text{AP}^{bb}_{M}$ metrics, while using $2.2\text{M}$ fewer parameters. When the Mask R-CNN detector is used, KCLA achieves overall performance comparable to existing linear layer attention methods. However, it demonstrates a notable lead of 0.8\% in metrics such as $AP_{L}^{bb}$.

\subsection{Medical image segmentation}
\textbf{Dataset.} 
We conduct the experiments on the ISIC2017 \cite{8363547}, ISIC2018 \cite{codella2019skin}, and Kvasir-SEG \cite{jha2020kvasir} datasets. The ISIC2017 dataset, released by the International Skin Imaging Collaboration (ISIC), constitutes a large-scale collection of dermoscopic images. This dataset contains 2,000 training images, 150 validation images, and 600 test images, each accompanied by corresponding ground truth segmentation masks. The ISIC2018 dataset expands upon this collection, comprising 2,594 training images alongside validation and test sets containing 100 and 1,000 images respectively. Kvasir-SEG \citep{jha2020kvasir} is a relatively small dataset, comprising 1,000 gastrointestinal polyp images with corresponding segmentation masks. All datasets was randomly partitioned into training and validation sets using a 7:3 split ratio. Images were normalized to $256\!\times\!256$ to maintain consistency.

\textbf{Settings.} 
We leveraged the training infrastructure, including training scripts and configuration file management, from the EGE-UNet codebase \citep{ruan2023ege}. The batch size is set to 8. The AdamW optimizer \citep{loshchilov2017decoupled} is employed with a CosineAnnealingLR scheduler \citep{loshchilov2016sgdr}, initialized with a learning rate of 1e-3 and weight decay of 1e-2. The scheduler was configured with a maximum of 50 iterations per cycle across 300 total training epochs, maintaining a minimum learning rate of 1e-5. We use ResNet-18 as the encoder of Unet to obtain ResNet-UNet and take it as the baseline model. We reduce the channel dimensions of each stage to 32, 64, 128, and 128 respectively, and adopt the same head dimension of 16 in all stages. All experiments were repeated five times, and the mean and standard deviation were reported. In the medical image segmentation experiment, due to the relatively small number of channels, the number of heads in each stage is also very small. Therefore, we removed the linear layer in the head attention part of KCLA to avoid overemphasizing certain parts. For non-layer attention methods, we directly obtained data from existing work \citep{ruan2023ege}. We apply different layer attention methods on the encoder of the backbone mentioned in Section~\ref{sec:experiments}, and report the mean and standard deviation of Mean Intersection over Union (mIoU) and Dice Similarity Coefficient (DSC), which are widely used for robust segmentation evaluation.

\textbf{Comparative results of medical image segmentation.}
The experimental results are presented in Table~\ref{tab:Results on ISIC2017 and ISIC2018}. Models identified with the suffix ``-UNet" are based on the baseline architecture detailed in the Settings section. The column labeled ``Layer attention" indicates whether the model incorporates a layer attention mechanism. Due to the adoption of a relatively smaller backbone and a reduced number of channels in experiment, the parametric cost of DLA-L was also significantly reduced, exceeding that of MRLA-L by only $0.03\text{M}$ parameters. Concurrently, the performance advantage of DLA-L completely vanished, resulting in an inferior performance compared to MRLA-L on both the ISIC2018 and Kvasir-SEG datasets. In contrast, KCLA consistently demonstrated superior performance over MRLA-L across all three datasets, while also significantly outperforming the baseline ResNet-UNet model.
\begin{table}[ht]
    \centering
    \caption{Ablation experiments on the main modules and settings in KCLA.}
    \label{tab:Ablation}
    \begin{tabular}{lclc}
    \toprule
       Model&Top1(\%)&Model&Top1(\%)\\
       \midrule
       Base& $75.45\pm0.30$&\textbf{(a)}-same $\mathcal{D}_i$&$75.23\pm0.09$\\
       \textbf{(b)}-w/o Exp&$75.33\pm0.35$&\textbf{(c)}-w/o HA&$75.25\pm0.27$\\
       \textbf{(d)}-RLA&$75.21\pm0.24$&\textbf{(e)}-h=2&$75.82\pm0.17$\\ 
 \textbf{(e)}-h=4& $75.66\pm0.21$& \textbf{(e)}-h=8& $75.57\pm0.23$\\
 \textbf{(e)}-h=32& $75.27\pm0.11$& \textbf{(f)}-w/o cs& $75.24\pm0.21$\\
 \bottomrule
    \end{tabular}
\end{table}

\begin{figure}
    \centering 
    \subfloat[FLOPS across network depths.]{%
        \includegraphics[width=0.49\linewidth]{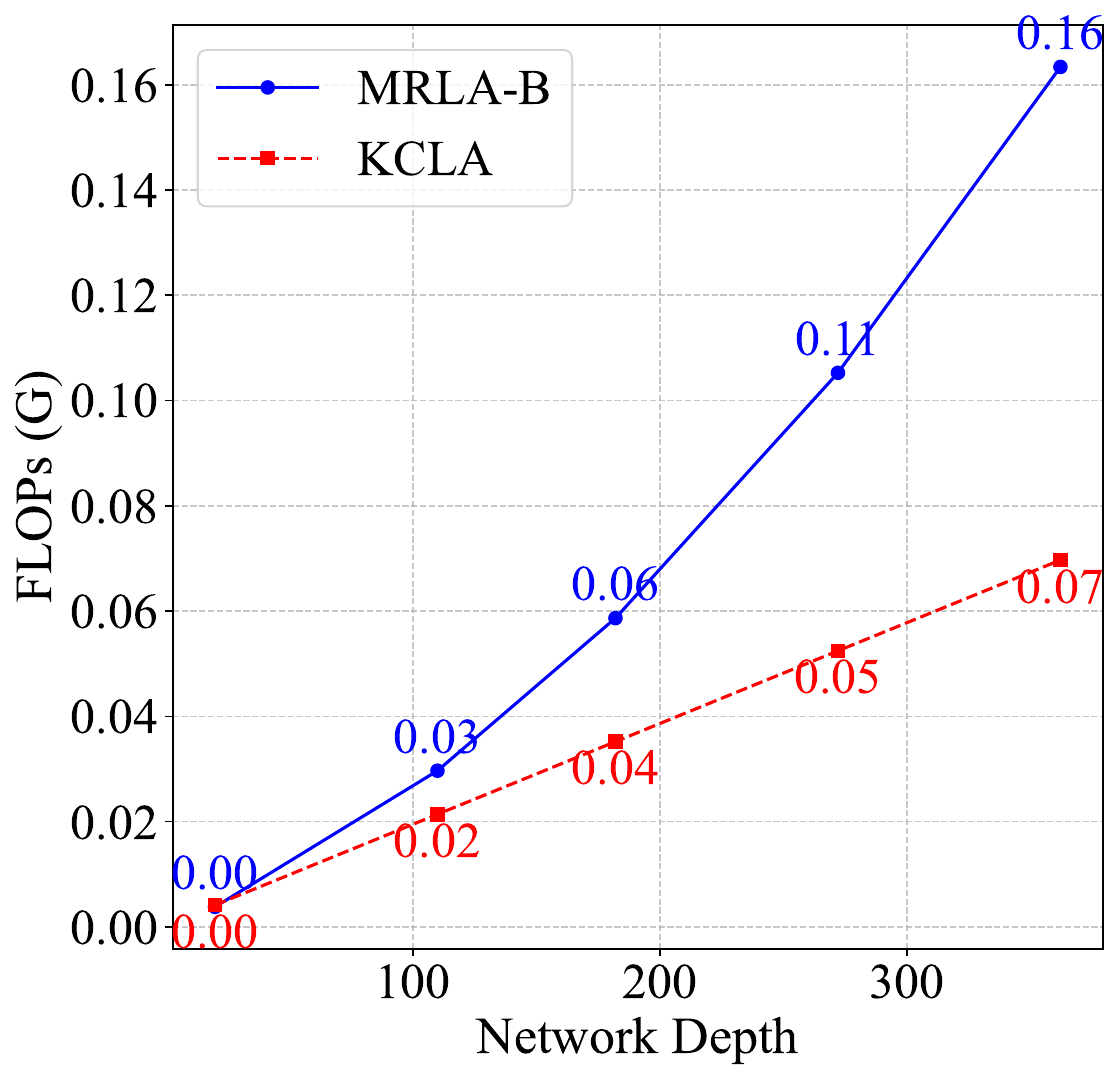}%
        \label{fig:flops comparison}
    }
    \subfloat[Memory across network depths.]{%
        \includegraphics[width=0.49\linewidth]{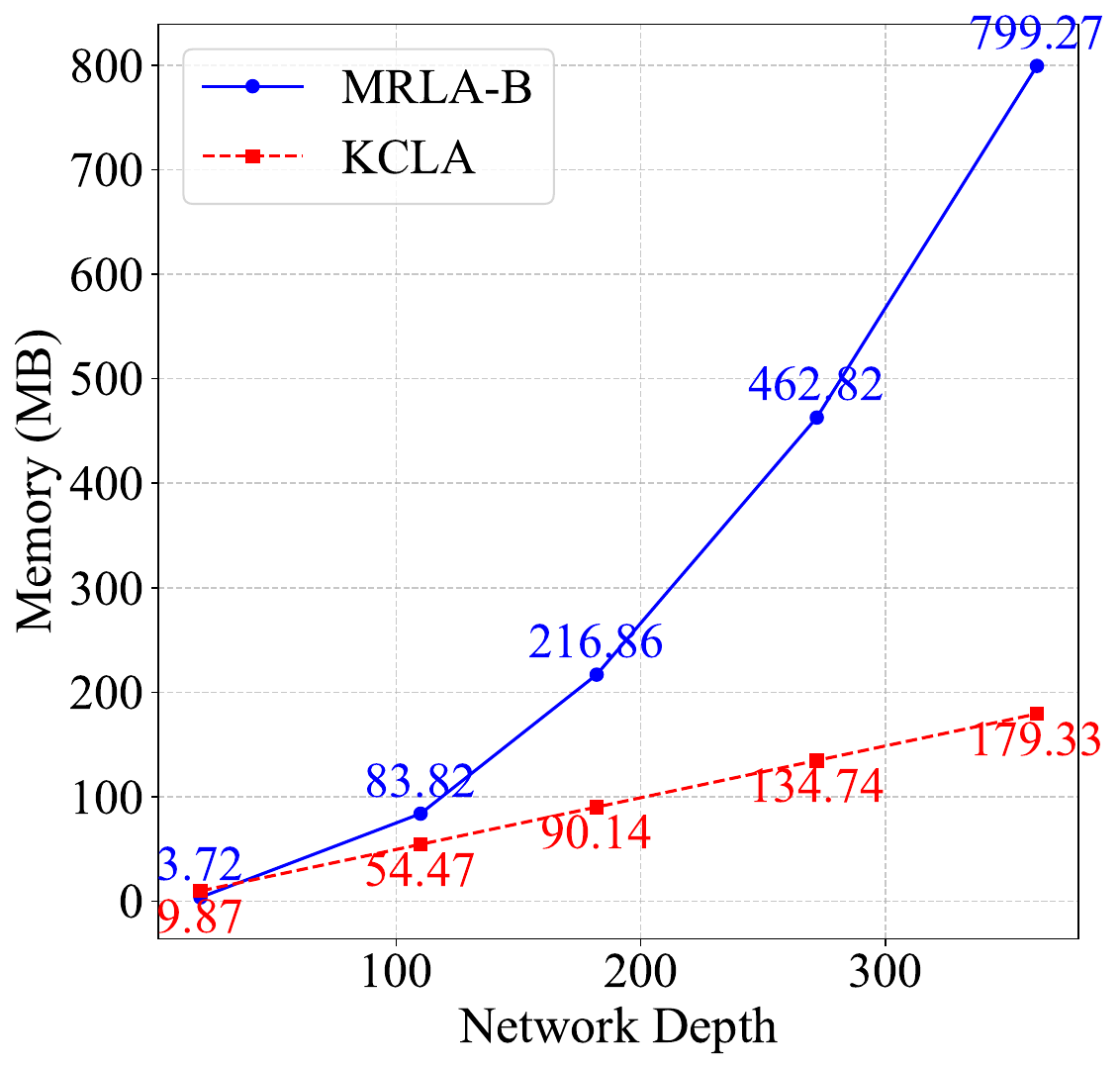}%
        \label{fig:memory comparison}
    }\\
    \caption{Comparison of the computational load and memory cost introduced by MRLA-B and KCLA as the depth increases.}
    \label{fig:memory and flops comparison}
\end{figure}

\subsection{Model complexity}
We quantified the additional overhead introduced by the MRLA-B and KCLA modules by calculating their respective computational complexity and memory cost, and subsequently subtracting the corresponding metrics of the backbone network. As shown in Fig.~\ref{fig:memory and flops comparison}, the incorporation of the attention mechanism causes the complexity and memory requirements of MRLA-B to grow quadratically with increasing network depth. In contrast, the overhead of KCLA exhibits a more favorable linear complexity growth for both metrics. It is noteworthy that other methods for linearizing general attention mechanisms require storing the matrices $\boldsymbol{}{K}$ and $\boldsymbol{}{V}$, thus exhibiting the same quadratic memory complexity as MRLA-B.

\begin{figure*}[t]
    \centering
    \subfloat[\label{fig:110-head1-attention-score}]{
        \includegraphics[width=\textwidth]{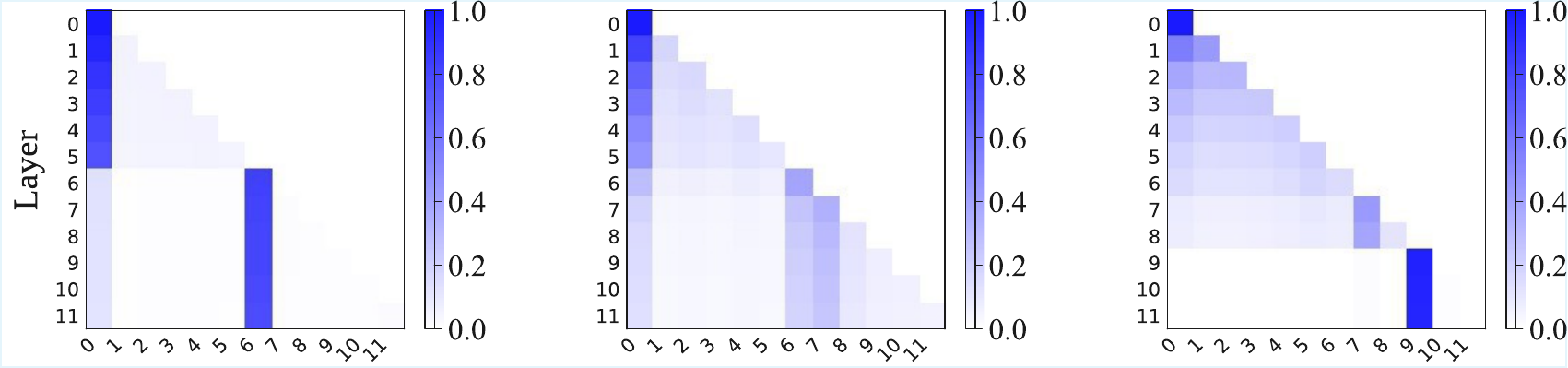}
    }
    \hfill
    \subfloat[\label{fig:110-head3-attention-score}]{
        \includegraphics[width=\textwidth]{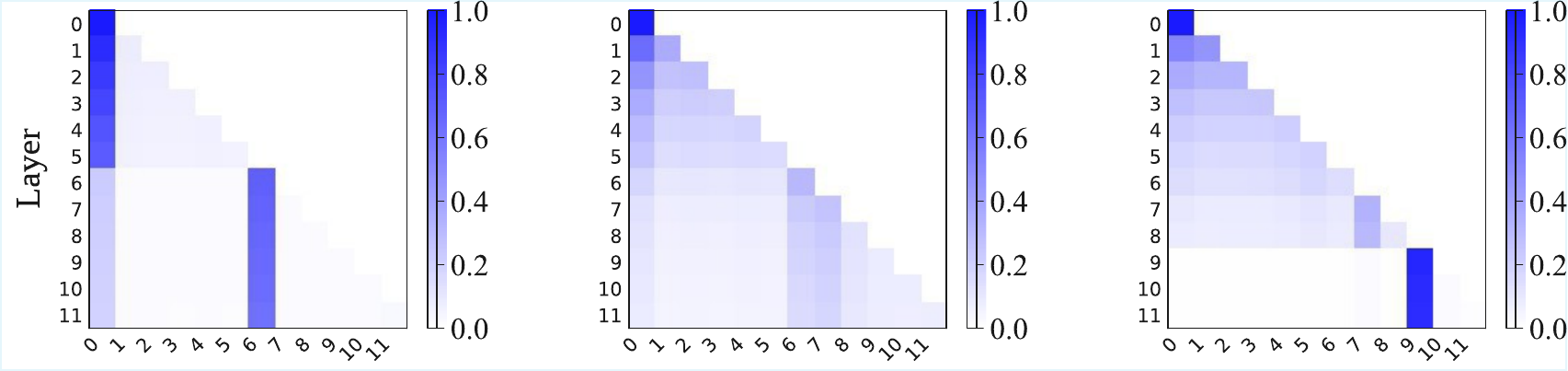}
    }
    \hfill
    \subfloat[\label{fig:110-head2-attention-score}]{
        \includegraphics[width=\textwidth]{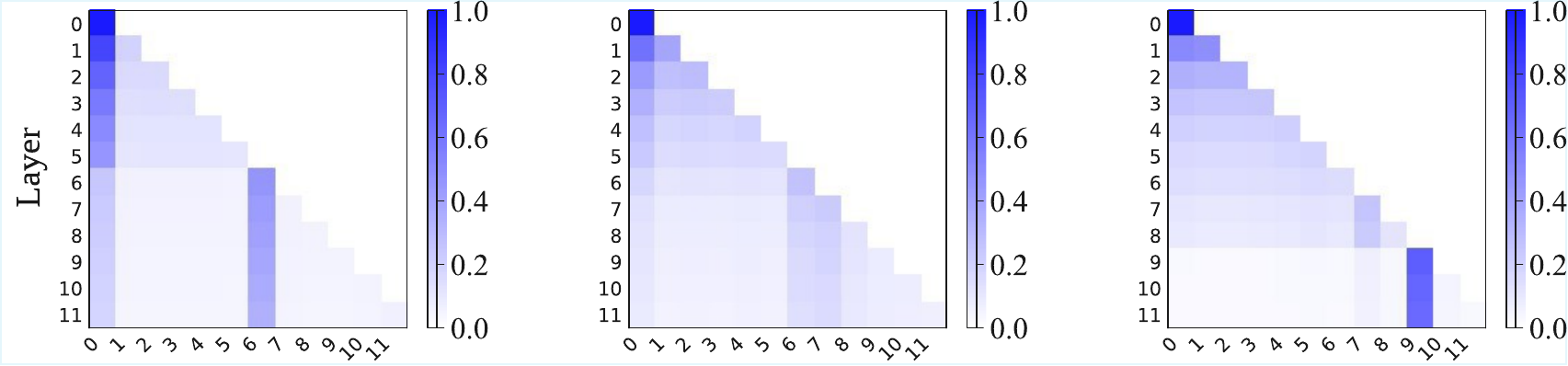}
    }

    \caption{The distribution of scores for different heads in KCLA-110. In this visualization, each row of small squares represents the attention distribution results, which sum up to one. (a) represents the first head, which exhibits the smallest scaling factor and the most concentrated attention distribution. (c) represents the final head, which has the largest scaling factor and consequently a more dispersed attention distribution. (b) shows an intermediate head. For each head, the distributions from left to right correspond to stage 1, stage 2, and stage 3, respectively.}
    \label{fig:kscore-resnet110}
\end{figure*}

\subsection{The distribution of KCLA attention scores for different heads}
\label{sec:kcla score on different head}
Since we use a modified multi-head attention mechanism, where each head uses a different scaling factor $\mathcal{D}_i$, with $i\in [1,H]$. The purpose of this design is to endow the heads of the model with different temperatures. Subsequently, by calculating the head attention scores for different heads and weighting them, the effect of dynamically adjusting the layer attention distribution is achieved. 

To observe the impact of different scaling factors on the distribution of attention scores, first, we calculate the attention score assigned by the $t$-th layer of the $h$-th head in KCLA to the $j$-th layer in the following way, where $j < t$:
\begin{equation}
\left\{
\begin{aligned}
\text{softmax}\left(\left[\|\boldsymbol{K}^t_{h[1,:]}\|,\ldots,\|\boldsymbol{K}^t_{h[t,:]}\|\right]\right)_j \\= \frac{\exp\left(\frac{\|\boldsymbol{K}^t_{h[j,:]}\|}{\mathcal{D}_h}\right)}{\sum_{i=1}^t \exp\left( \frac{\|\boldsymbol{K}^t_{h[i,:]}\|}{\mathcal{D}_h}\right)}
\end{aligned}
\label{eq:softmax normalized kcla score}
\right.
\end{equation}
We separately calculated the attention scores of each layer in the first head, the middle head, and the last head in ResNet-50 and ResNet-110 for the previous layers. Among them, the scaling factor of the first head is the smallest, the middle head adopts the standard scaling factor, and the scaling factor of the last head is the largest. The results are shown in Fig.~\ref{fig:kscore-resnet110}. It can be observed that even when the last layer of each stage computes attention, a certain weight may still be assigned to earlier layers. In contrast, the contribution of earlier layers in MRLA-L is exponentially attenuated, effectively vanishing. Although different heads produce similar distributions, there are still differences in their degree of concentration. When the scaling factor is small, a concentrated attention distribution is generated, as shown in Fig.~\ref{fig:110-head1-attention-score}. When the scaling factor is large, the attention distribution is more uniform (Fig.~\ref{fig:110-head2-attention-score}). 

\section{Ablation study}
We conducted ablation studies using ResNet-56 on the CIFAR-100 dataset to evaluate different variants of KCLA. The baseline KCLA uses $h$ = 16. Here $h$ is the dimension of a single head. We systematically examined several configurations, including \textbf{(a)} setting all elements in $\mathcal{D}$ to $\sqrt{h}$; \textbf{(b)} removing the exponential transformation applied to the norms of Key representations; \textbf{(c)} excluding the head attention module from Equation~(\ref{eq:KCLA_multihead_output}); \textbf{(d)} applying KCLA to linearized RLA; \textbf{(e)} varying the head dimension; and \textbf{(f)} disabling cross-stage design. Each experiment was repeated five times and then calculate the mean and standard deviation. The experimental results, presented in Table~\ref{tab:Ablation}, demonstrate the significant contribution of each component to the model's performance. When the key designs in KCLA, including different $\mathcal{D}_i$, exponential transformation, head attention, and cross-stage design, are removed, the model accuracy drops by 0.12\% to 0.22\%. When the RLA form is adopted, the accuracy drops by 0.24\%, demonstrating the ability of KCLA to preserve the dynamics in the standard layer attention. Furthermore, the dimension of attention heads exhibits considerable influence on KCLA's effectiveness. Reducing the head dimension results in improved model accuracy, but it also leads to the head attention module introducing more parameters. For example, when h = 2, it will introduce an additional 0.1M parameters.

\section{Conclusion}
\label{sec:conclusion}
This paper designs and proposes a novel linear layer attention method. By directly and accurately approximating the standard layer attention mechanism, our method avoids the static and rigid feature representation issues inherent in RLA. Crucially, compared with MRLA-L, our approach maintains the important capability for long-distance dependency modeling. This retained capability allows the model to efficiently aggregate global context information while successfully mitigating the quadratic computational complexity typically associated with standard attention, ensuring a desirable balance between efficiency and performance. We conducted comprehensive evaluations of the proposed method across several core computer vision benchmarks. The experimental results consistently demonstrate that in the three key tasks of image classification, object detection, and medical image segmentation, our method has achieved performance close to or even exceeding those of previous SOTA models.  Furthermore, the method of restoring the softmax function introduced in this paper was shown to be a critical factor for enhancing model performance. However, one aspect remains open for future optimization: our current implementation relies solely on pre-set temperature parameters. Given that recent research has explored learnable dynamic temperature parameters \cite{oorloff2025mitigating}, offering finer control and adaptability, our future work will therefore primarily focus on designing a more advanced, self-adaptive, and dynamic temperature multi-head linear layer attention method.

\bibliography{bib2026}
\bibliographystyle{IEEEtran}


 





\end{document}